\pdfoutput=1

\documentclass[11pt]{article}

\usepackage{acl}

\usepackage{times}
\usepackage{latexsym}

\usepackage[T1]{fontenc}

\usepackage[utf8]{inputenc}

\usepackage{microtype}

\usepackage{graphicx}
\usepackage{subfiles}
\usepackage{multirow}
\usepackage{booktabs}
\usepackage{tabularx}

\title{Crowdsourcing on Sensitive Data with Privacy-Preserving Text Rewriting}

\author{Nina Mouhammad$^{a}$, Johannes Daxenberger$^{b}$, Benjamin Schiller$^{b}$ \and Ivan Habernal$^c$ \\
    $^a$DIPF, Leibniz Institute for Research and Information in Education, \\
    $^b$summetix GmbH,
    $^c$ Trustworthy Human Language Technologies, \\ Department of Computer Science, Technical University of Darmstadt \\ 
    $^a$\texttt{n.mouhammad@dipf.de}, $^b$\texttt{\{schiller,daxenberger\}@summetix.com},\\$^c$
    \texttt{ivan.habernal@tu-darmstadt.de}}

\begin{document}
\maketitle
\begin{abstract}
Most tasks in NLP require labeled data. Data labeling is often done on crowdsourcing platforms due to scalability reasons. However, publishing data on public platforms can only be done if no privacy-relevant information is included. Textual data often contains sensitive information like person names or locations. In this work, we investigate how removing personally identifiable information (PII) as well as applying differential privacy (DP) rewriting can enable text with privacy-relevant information to be used for crowdsourcing. We find that DP-rewriting before crowdsourcing can preserve privacy while still leading to good label quality for certain tasks and data. PII-removal led to good label quality in all examined tasks, however, there are no privacy guarantees given.
\end{abstract}

\section{Introduction}

For supervised NLP tasks, large amounts of labeled data are needed. In many cases, only unlabeled data is available and labeling is then performed via crowdsourcing/crowdworking platforms like Amazon Mechanical Turk (AMT). These crowdworking platforms are used because they provide a time-efficient way to obtain labels for unlabeled data, making the annotation task easily scalable.

However, data should only be published on crowdsourcing platforms if it contains no privacy-relevant information. Unfortunately, it is not always obvious what is privacy relevant and what is not \citep{Narayanan12}. As a consequence, most textual datasets cannot be annotated on crowdworking platforms if the privacy of affected persons contained in the data needs to be respected.

A common practice is to automatically replace personally identifiable information (PII) in a text. However, not all privacy-relevant information is contained in PII \citep{Narayanan12} and the automatic detection of PII does not work perfectly. Therefore, PII-removal alone is no guarantee that privacy is preserved. 

An approach that can actually give privacy guarantees is differential privacy (DP). DP offers formal mathematical guarantees for privacy-preserving data publishing, which has most recently also been applied to textual data \cite{DPText, Krishna21, Bo21}. The benefit of using differential privacy is that it is possible to set an upper boundary for privacy risks. Therefore, one exactly knows how large the privacy risk is and can set it to a sufficiently low level when using DP.


In this work, we want to explore different privacy preservation techniques for textual data in the context of crowdsourcing. 
We do this by performing crowdsourcing on data which has been modified by using DP rewriting, PII-removal, or a combination of both.
We show that there is a tradeoff between privacy and utility (label quality) when deciding for one of these methods, how this tradeoff is expressed and how it depends on the chosen task and data. Furthermore, we provide recommendations which task properties might lead to the most desirable results.


\section{Related work}
Privacy leakages can have harmful consequences for individuals. Therefore, privacy protection is regulated by law in some parts of the world, e.g., by the GDPR in Europe \citep{GDPR16} or the HIPAA Act \citep{hipaa} for medical data in the US. Unfortunately, it is impossible to fully prevent the risk of privacy leakages. Therefore, the ultimate goal is to reduce this risk.

A common practice to reduce the risk of privacy leakages in textual data is to automatically detect and replace personally identifiable information (e.g. \citealp{Ge20, Pilan22, Eder20}). This approach is called PII-removal in the following. However, there are two problems with PII-removal. First, without PII-labeled training data, in most cases named entity recognition or regular expressions are used for PII-removal \citep{Ge20, Pilan22, Eder20}. This narrows down which kind of PII can be detected. Second, there is no possibility to quantify the remaining privacy risk.

Differential privacy (DP) solves the problem of estimating privacy risks. It is a mathematical concept, supposed to enable sharing datasets
containing private information without giving away this private information \cite{Dwork14}. It has recently been applied in NLP for rewriting texts in a differentially private way \cite{Krishna21, Bo21, DPText}. The basic idea of `local' differential privacy rewriting for textual data is to add noise to each data point. As a result, the probability of distinguishing data belonging to one individual from data of any other individual in the dataset is bounded. Furthermore, it is possible to define how small the probability of being able to distinguish this data should be by setting the `privacy budget' parameter called $\epsilon$.

\section{Data}
Three corpora were used for the experiments: ATIS \cite{Tur10}, SNIPS \cite{Coucke18} and TripAdvisor (TA) \cite{Li13}. 
The ATIS corpus consists of transcriptions of flight information requests and the task is to classify them based on their intent. There are different versions of the ATIS corpus available, we use it in the form provided by \citet{Tur10}. 
SNIPS \cite{Coucke18} is an intent classification dataset as well and consists of instructions for voice assistants.
TripAdvisor \cite{Li13} (TA) contains hotel reviews. We use only the titles of these hotel reviews because the full review texts were too long.

We chose those datasets based on multiple criteria. First, we had some task-specific criteria. The task should be relevant in real-world use cases, it should not require previous knowledge and it should be simple and quick to solve. Second, we had some text-specific criteria. The texts should contain privacy relevant information, it should be in clear and generally understood language and the text snippets should be short. Furthermore, all datasets should have high-quality gold labels so that we could compare the labels obtained in our experiments with these gold labels. Finally, these datasets have been used in related works on privacy text rewriting.

To simplify the task further, we reduced all of them to binary labelling tasks. This means that we chose one class as target class (e.g. ``Airfare'' for ATIS) and defined the task as deciding whether a given data point belonged to that target class or not. Furthermore, we only included data points which consisted of less than 200 characters for the crowdsourcing, but still used the longer texts for the DP pretraining in order to have enough pretraining data. An overview of the properties of all corpora in the modified versions used in this work can be found in Table \ref{tab:dataset_stats}. Furthermore, example sentences are shown in Table \ref{tab:example_sentences}.


\begin{table}[]
    \centering
    \begin{tabular}{c|r|r|r|r} \toprule
        \multirow{2}{*}{corpus} & \multicolumn{2}{c|}{data points}  & \multicolumn{2}{c}{avg. length} \\\cline{2-5}
        & target & rest & target & rest \\
        \midrule 
         ATIS & 403 & 4100 & 67.91 & 66.77\\ 
        \hline 
         SNIPS & 1936 & 11681 & 48.24 & 46.33\\ 
        \hline 
         TA & 19663 & 9974 & 181.48 & 298.96\\ \bottomrule
    \end{tabular}
    \caption{Number of data points (``data points'') and average number of characters per data point (``avg length'' per corpus in our modified version of the corpora. ``target'' stands for ``target class'' and ``rest"'' for all data points not belonging to the target class.}
    \label{tab:dataset_stats}
\end{table}

\begin{table*}[ht]
    \centering
    \begin{tabularx}{\textwidth}{c|X|X} \toprule
         & \bf{target class} & \bf{not target class}  \\ \midrule
         \multirow{2}{*}{ATIS} & cheapest airfare from tacoma to orlando & what flights are available from pitsburgh to baltimore on thursday morning \\
         & show me all the one way fares from tacoma to montreal & what is the arrival time in san francisco for the 755 am flight leaving washington? \\ \hline
         \multirow{2}{*}{SNIPS} & add The Crowd to corinne's acoustic soul playlist & Book a restaurant in El Salvador for 10 people. \\
         &  add this track to krystal's piano 100 & Play a chant by Mj Cole \\ \hline
         \multirow{2}{*}{TA} & AMAZING Concierge Staff/Eric Sofield is the best & Avoid lower floors... especially room 202 \\
         & Best Hotel in Philly & Bugs and terrible housekeeping \\ \bottomrule
         
    \end{tabularx}
    \caption{Examples per corpus and class.}
    \label{tab:example_sentences}
\end{table*}

\section{Model}
\paragraph{PII-removal}
The PII-removal is based on regular expressions and on spacy \cite{Honnibal20} which we used for named entity recognition and part of speech tagging. With spacy, we detected names of persons, locations,  dates and times. Those were then replaced with the strings "<NAME>", "<LOCATION>", "<DATE>" and "<TIME>". Additionally, we used regular expressions, to replace other personal information like mail addresses and phone numbers.

\paragraph{DP-rewriting}
For DP-rewriting we used the work of \citet{DPText}. They provide an open-source framework for DP rewriting with a trainable model based on the idea behind ADePT \cite{Krishna21}. This model consists of an auto-encoder which is pretrained first to learn how to compress texts. Afterwards, the texts to be rewritten are transformed into a compressed version, noise according to either a Gaussian or Laplacian distribution is added and then the text is reconstructed based on this vector. We used Gaussian noise and set $\delta = 1*10^{-4}$. For $\epsilon$, different values were used in different experiments. We state which value has been used when explaining each of the experiments. Furthermore, we did not append the class labels (as proposed in \cite{Krishna21}), because usually class labels are only crowdsourced if there are none yet. 

For each corpus, we split the data into three different subsets, one for pretraining, one for validation of the pretraining and one that will be rewritten for the crowdsourcing. Based on this, we created six differently pretrained models. For each corpus, we had one model pretrained with the unchanged pretraining data and one pretrained with the pretraining data after PII were replaced. 

\paragraph{Rewriting pipelines}
\begin{figure}[ht]
    \centering
    \includegraphics[width=\columnwidth]{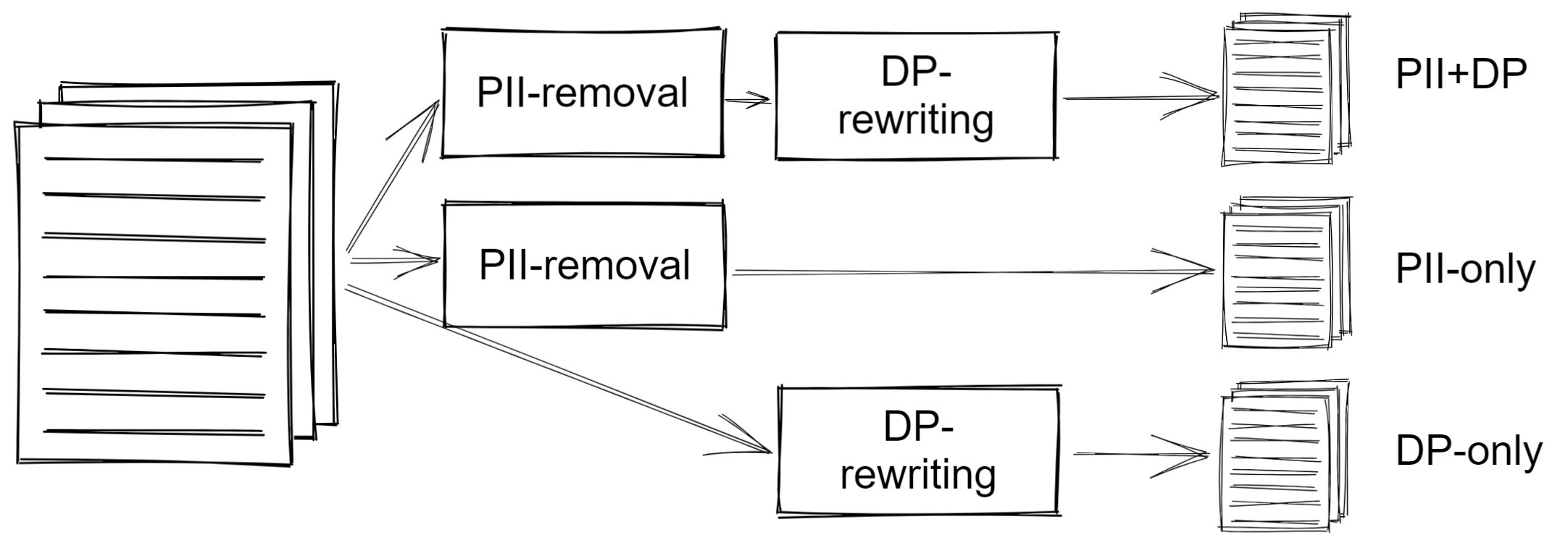}
    \caption{We used three different rewriting pipelines: PII-only, DP-only and PII + DP. They are depicted here.}
    \label{fig:experiment_pipelines}
\end{figure}

We created three different rewriting pipelines so that we can compare the two chosen rewriting methods and the combination of them. For each rewriting method, there is one pipeline where only this rewriting method is applied to privatize the data (PII-only and DP-only). Furthermore, there is one pipeline where we first perform PII-removal and then DP-rewriting (PII + DP). They are visualized in Figure~\ref{fig:experiment_pipelines}. After the data has been rewritten in different ways, we requested annotations based on our binary labeling task on Amazon Mechanical Turk. An example HIT can be found in the Appendix~\ref{sec:example_HIT}. All crowdworkers were from the US. Therefore, the payment per HIT was calculated based on the US minimal wage in order to guarantee fair payment.

\section{Results}

\paragraph{PII-only vs. DP-only vs. PII + DP}

First, we wanted to explore general differences between the three rewriting pipelines. Therefore, we run the data through all pipelines and requested annotations from 5 crowdworkers per pipeline and data point. For the DP-rewriting in DP-only and PII + DP we set $\epsilon = 10000$. This is a very high choice for $\epsilon$. However, it was the smallest value which ensured that the resulting text still had some very basic utility.

After the annotation, we aggregated the individual annotations per data point by using MACE \cite{Hovy13} with a threshold of 1. Then we compared these aggregated labels to the original gold labels by calculating F1-scores (see Table \ref{tab:f1-scores_pipeline_comparison}).

\begin{table}[ht]
    \centering
    \begin{tabular}{c|r|r|r} \toprule
         Pipeline & ATIS & SNIPS & TA  \\ \midrule
         PII + DP & 0.377 & 0.828 & 0.588 \\
         DP-only & 0.549 & 0.935 & 0.698\\
         PII-only & \textbf{0.949} & \textbf{0.991} & \textbf{0.932} \\ \bottomrule
    \end{tabular}
    \caption{F1-scores of the original gold labels compared to the labels obtained in our experiments. The highest value per column is indicated in bold. Differences per row were statistically significant with $\alpha = 0.05$ for all values.}
    \label{tab:f1-scores_pipeline_comparison}
\end{table}

PII-only performed best for all corpora regarding the F1-score. Furthermore, DP-only led to better F1-scores than PII + DP. However, this depicts only the performance regarding gold label quality. Regarding privacy, it is the other way around. This will be discussed in more detail in Section~\ref{sec:Discussion}.

Apart from this, in Table \ref{tab:f1-scores_pipeline_comparison} we can see that there are differences between the corpora, especially regarding DP-rewriting. For the SNIPS corpus, the DP-rewriting had a far smaller negative effect on the F1-scores than on the TA corpus or even the ATIS corpus.

\paragraph{The effect of $\epsilon$}
In DP-rewriting, the $\epsilon$-parameter is the most important parameter, because it represents the privacy guarantee. A high value stands for high privacy risks. To investigate the effects of this $\epsilon$-parameter, we reran the DP-only pipeline in a slightly modified way. We set $\epsilon = 3333$ and requested annotations from three different crowdworkers per pipeline and data point. Then, again, we aggregated the annotations per pipeline and data point by using MACE \cite{Hovy13} and calculated the F1-scores in comparison to the original gold labels. 

\begin{table}[]
    \centering
    \begin{tabular}{c|r|r} \toprule
        Corpus & $\epsilon = 3333$ & $\epsilon = 10 000$   \\ \toprule
        ATIS & 0.229 & \textbf{0.517} \\
        SNIPS & 0.519 & \textbf{0.920} \\
        TA & 0.426 & \textbf{0.687} \\ \bottomrule
    \end{tabular}
    \caption{F1-scores of the same data rewritten with DP-only and different values for $\epsilon$. Differences per row are statistically significant with alpha = 0.05. The highest value per row is highlighted in bold.}
    \label{tab:epsilon_comparison}
\end{table}

We compared the F1-scores to the F1-scores of the data rewritten with $\epsilon = 10000$. To guarantee a fair comparison, we only used 3 annotations per data point as well and reaggregated them with MACE (see Table~\ref{tab:epsilon_comparison}).
For all corpora, the lower $\epsilon$ resulted in statistically significantly lower F1-scores. With the lower $\epsilon$, the performance difference between SNIPS and the other corpora decreased.

\paragraph{Multiple rewritten versions}
While lower $\epsilon$ values increase privacy, they decrease the utility drastically. But what if we rewrite multiple times with the same $\epsilon$, but different random seeds and then aggregate the crowdsourced annotations? Can the differently added noise be counterbalanced by this so that utility is overall increased?

\begin{figure}[ht]
    \centering
    \includegraphics[width=\columnwidth]{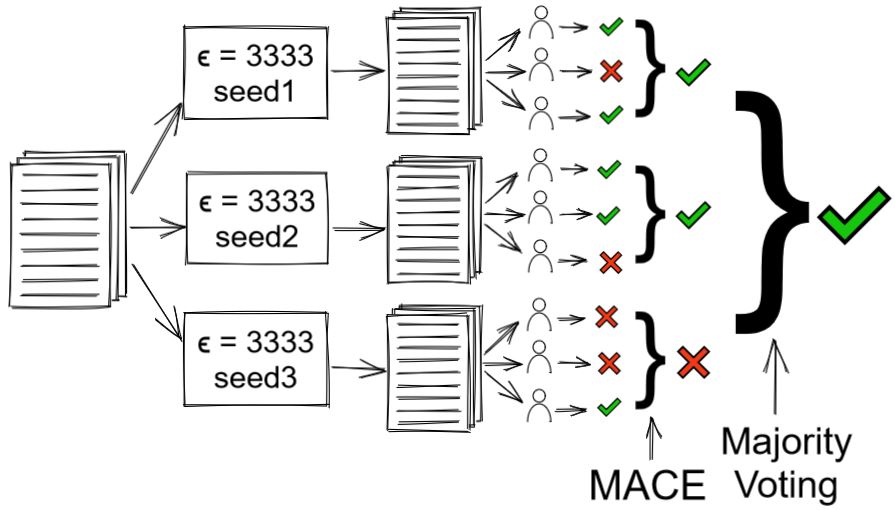}
    \caption{Process of generating multiple differently rewritten versions and aggregating their annotations.}
    \label{fig:multiple_rewritten_versions}
\end{figure}

For each data point, we created two other versions rewritten with DP-only and $\epsilon = 3333$. Then we requested three annotations per version from crowdworkers and aggregated the annotations per data point over all versions. 
This time, we could not use MACE \cite{Hovy13} to aggregate the data, because for using MACE the annotations need to be independent when conditioned on the true labels. However, in our case, they are only independent when conditioned on the true labels and the corresponding rewritten version. Therefore, we could only use MACE to aggregate the annotations per version and aggregated the results of this by using majority voting. The whole process is illustrated in Figure~\ref{fig:multiple_rewritten_versions}.

Again, we calculated F1-scores between our aggregated labels and the original gold labels. The results, as well as a comparison to the previous results, can be found in Table~\ref{tab:multiple_rewritten_versions}. Interestingly, using multiple differently rewritten versions did not increase, but decreased the F1-scores for all corpora except SNIPS.

\begin{table}[]
    \centering
    \begin{tabular}{c|r|r|r} \toprule
        \multirow{2}{*}{Corpus} & \multirow{2}{*}{$\epsilon = 3333$} &  multiple & \multirow{2}{*}{$\epsilon = 10 000$}   \\ 
        & & versions & \\ \midrule
        ATIS & 0.229 & 0.180 & \textbf{0.517} \\
        SNIPS & 0.519 & 0.519 & \textbf{0.920} \\
        TA & 0.426 & 0.350 & \textbf{0.687} \\ \bottomrule
    \end{tabular}
    \caption{F1-scores of the same data rewritten with DP-only and different values for $\epsilon$. The highest value per row is highlighted in bold.}
    \label{tab:multiple_rewritten_versions}
\end{table}

We explored different aggregation methods. They can be divided into two types: two-step-aggregation and one-step-aggregation. The two-step-aggregation methods consist of two steps: In the first, there is an aggregation per rewritten version and in the second step, these aggregations are aggregated again. The aggregation we used for Table~\ref{tab:multiple_rewritten_versions} and illustrated in Figure~\ref{fig:multiple_rewritten_versions} is a two-step aggregation method with MACE as the first step and majority voting as the second step. In the one-step-aggregation methods, all annotations of all versions are aggregated in one single step with one aggregation technique.

The aggregation methods were chosen based on commonly occurring problems in our experiments. In general, it was very noticeable, that there were far more cases where data points that belong to the target class were not recognized as belonging to the target class than the other way around. Therefore, we created a threshold-based aggregation method for this. It is a one-step-aggregation method and the idea is, that the target class is chosen if more than x annotations of one data point are target class annotations. So if we have a threshold of x = 3 and a data point with four target class annotations and five non-target class annotations, the aggregated label will be the target class label. If there were only three target class annotations and six non-target class annotations, the aggregated label would be the non-target class annotation. This method will be abbreviated as tx in the following, where x is replaced with the used threshold.

Based on that threshold idea, we also created a two-step-aggregation method where first, annotations per version were aggregated with MACE and afterwards the aggregated labels were aggregated with a threshold of 0. This method will be abbreviated as MACE\_t0. Furthermore, we tried plain majority voting in a one-step-aggregation (MV), majority voting in a two-step-aggregation (MV\_MV) and the previously discussed two-step-aggregation with MACE and majority voting (MACE\_MV).

\begin{table}[ht]
    \centering
    \begin{tabular}{c|r|r|r}\toprule
        Aggregation & ATIS & SNIPS & TA  \\ \midrule
        MV & 0.050 & 0.297 & 0.260   \\
        t0 & \textbf{0.448} & \textbf{0.799} & \textbf{0.638}  \\
        t1 & 0.368 & 0.730 & 0.581  \\
        t2 & 0.322 & 0.648 & 0.503  \\
        \hline
        MV\_MV & 0.078 & 0.313 & 0.269  \\
        MACE\_MV & 0.180 & 0.519 & 0.350  \\
        MACE\_t0 & 0.431 & 0.777 & 0.604  \\ \bottomrule
    \end{tabular}
    \caption{Comparison of different aggregation methods for the annotations of multiple rewritten versions. The highest value per column is highlighted in bold.}
    \label{tab:different_aggregation_methods}
\end{table}

Per aggregation method, we calculated the F1-Scores of the resulting labels and the original gold labels (see Table~\ref{tab:different_aggregation_methods}). The methods which do not take into consideration that target class data points have been mislabeled more often than non-target class points give the worst results. The methods taking this point into consideration lead to a lot better F1-scores. The most extreme method, t0, in which a data point is labeled as target class if only one crowdworker annotated one version as target class, lead to the best F1-scores.

\section{Discussion} 
\label{sec:Discussion}
\paragraph{Corpus differences}

The negative effect on the utility of DP-rewriting in our experiments has been corpus dependent. In the following, we will explore reasons for this. 

As already discussed before, the lower F1-scores can mainly be traced back to data points which belong to the target class but have not been recognized as belonging to the target class. While this problem exists for all corpora, it is least prominent for SNIPS, see Table~\ref{tab:target_class_preservation}.

\begin{table}[]
    \centering
    \begin{tabular}{c|c|r} \toprule
        Corpus & Gold & DP-only  \\ \midrule
        ATIS & 29.41\% & 13.10\% \\
        SNIPS & 50.00\% & 42.64\% \\
        TA & 50.00\% & 36.86\% \\ \bottomrule
    \end{tabular}
    \caption{Percentage of data points in the crowdsourcing set labelled as target class according to the original gold labels (``Gold'') and according to the labels gained by crowdsourcing after using DP-only with $\epsilon = 10 000$ (``DP-only'').}
    \label{tab:target_class_preservation}
\end{table}

To explore potential reasons for the indifference of target class non-recognition, we will use a concept we call \emph{indicator words}. Indicator words are words which do not appear equally often in the target class and the non-target class data. For example, for ATIS the target class is ``Airfare'', meaning that all requests asking about prices for flights belong to that class. Words that therefore often occur in the target class, but not in the non-target class data are ``fare'', ``airfare'', ``cost'', etc. While it is not possible to correctly identify the class based on only these indicator words, they are helpful signals in many cases and therefore a useful approximation to explore the indifference in the class recognition further.

For ATIS and TA, the usefulness of indicator words has been substantially decreased by the DP-rewriting, as we can see in Table~\ref{tab:indicator_words_DP-only}. Based on the given tasks, indicator words indicate the affiliation to the target class (like in ATIS and SNIPS) or the affiliation to the non-target class (like in TA). After DP-rewriting, we see that in ATIS the target class indicator words occurred only half as often in target class texts as before, while this was not the case in non-target class texts. In TA, the non-target class indicator words appeared less often in the non-target class texts but more often in the target class texts than before. In both cases, the difference between the target class and the non-target class, as approximated by indicator words has been decreased. For SNIPS, however, no such clear effect could be observed.

\begin{table}[]
    \centering
    \begin{tabular}{c|c|r|r} \toprule
        Corpus & Version & Target & Rest  \\ \midrule
        \multirow{2}{*}{ATIS} & original & 232 & 21\\
        & DP-only & 104 & 24 \\ \hline
        \multirow{2}{*}{SNIPS} & original & 520 & 2 \\
        & DP-only & 596 & 6 \\ \hline
        \multirow{2}{*}{TA} & original & 5 & 142 \\
        & DP-only & 48 & 118 \\ \bottomrule
    \end{tabular}
    \caption{Distribution of indicator words for the target class (ATIS and SNIPS) or the non target class (TA) before and after DP-only.}
    \label{tab:indicator_words_DP-only}
\end{table}

This assimilation of both classes according to the indicator words in ATIS and TA, but not in SNIPS is due to the relative uncommonness of these indicator words. The basic idea of the version of DP we use is that uncommonness in the dataset is correlated with the probability of being removed. Therefore, uncommon words have a higher probability of being removed than common words. For SNIPS, we had only two indicator words and they occurred 522 times in the original dataset. For ATIS, we had six different indicator words and all of them only occurred 253 times. This is even more extreme in TA, where we used basically all negatively connoted words as indicator words and nevertheless there were only 147 of them in the original corpus. This relative uncommonness of the indicator words in ATIS and TA is the reason why they have often been replaced during DP-rewriting.

However, based on this argumentation, the F1-score as well as the difference between the classes regarding the indicator words should have been higher for ATIS than for TA. Why is this not the case? It can probably be traced back to the pretraining data. For ATIS, the original dataset was very small and imbalanced. Therefore, only 4.28\% of the pretraining data (compared to 29.41\% of the crowdsourcing data) has been from the target class. This further reduced the uncommonness of the indicator words, especially in comparison to TA where 50\% of the pretraining data came from the target class.

\begin{figure}[ht]
        \centering
        \includegraphics[width=\columnwidth]{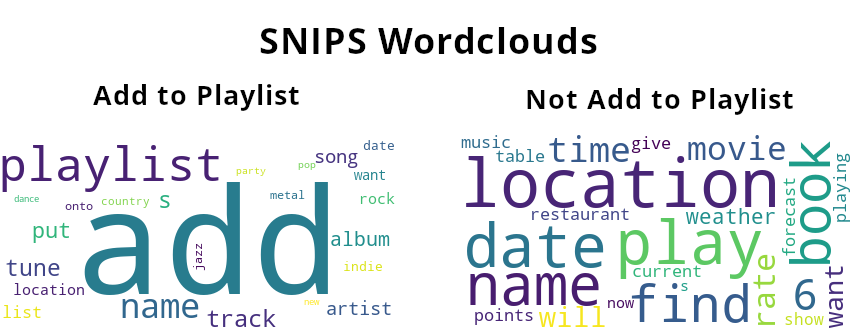}
        \caption{Wordcloud for the 25 most common non-stopword words per class of the PII-only version of SNIPS}
        \label{fig:wordcloud_SNIPS}
    \end{figure}

\begin{figure}[ht]
        \centering
        \includegraphics[width=\columnwidth]{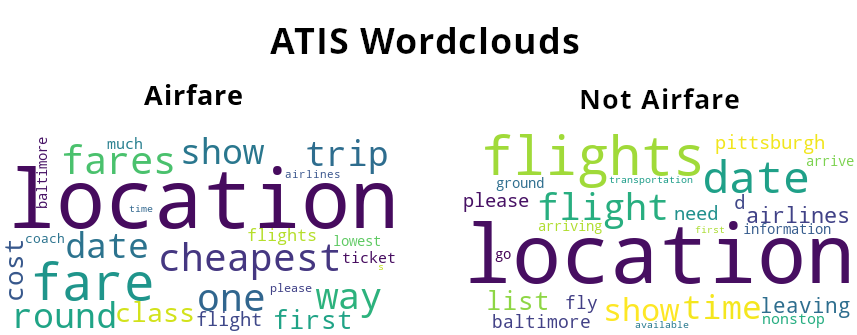}
        \caption{Wordcloud for the 25 most common non-stopword words per class of the PII-only version of ATIS}
        \label{fig:wordcloud_ATIS}
    \end{figure}

\begin{figure}[ht]
        \centering
        \includegraphics[width=\columnwidth]{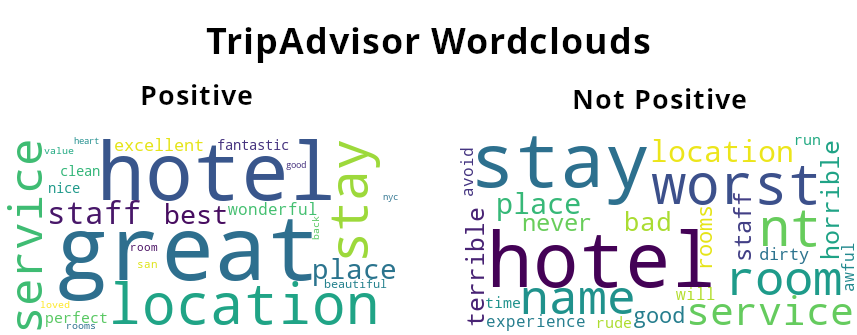}
        \caption{Wordcloud for the 25 most common non-stopword words per class of the PII-only version of TA}
        \label{fig:wordcloud_TA}
    \end{figure}

Another important factor is the amount of difference between the two classes. If the target class and the non-target class are very similar, changing one word might already change the class. If they are very different, a change of one word does not affect which class a text belongs to. To illustrate the differences between the two classes per corpus, we created wordclouds containing the 25 most common non-stopwords per class (see Figures~\ref{fig:wordcloud_SNIPS}, \ref{fig:wordcloud_ATIS}, \ref{fig:wordcloud_TA}). For this, we used the PII-only version of the datasets, because then e.g. locations were summarized by ``location'' and the wordclouds are easier to grasp.
            
Figure~\ref{fig:wordcloud_SNIPS} shows that the target class ``Add to Playlist'' of the SNIPS corpus is very different from the non-target class ``Not Add to Playlist''. Furthermore, the indicator words ``add'' and ``playlist'' are very prominent in the target class, but not in the non-target class. For ATIS, the wordclouds of the two classes are less different, see Figure~\ref{fig:wordcloud_ATIS}. Furthermore, in ATIS relatively small changes can cause a class change. The sentence ``How much is the cheapest flight from Pittsburgh to Baltimore?'' belongs to the class ``Airfare'', while ``What is the cheapest flight from Pittsburgh to Baltimore?'' does not belong to the class ``Airfare'' because the answer to this question would not be a price. There are many more examples like this in ATIS, but not in SNIPS.

For TA, the wordclouds are also less different than for SNIPS. Additionally, there are also cases where changing one word changes the whole class. For example ``Best hotel in Philly'' could be changed to ``Worst hotel in Philly'' and would then belong to the other class. However, there are fewer cases like this in TA than in ATIS.

All in all, there are multiple reasons explaining the corpus differences. First, the balance in the pretraining data is important, especially for very small corpora. Second, the diversity of the corpus, in relation to the corpus size affects the utility. And third, the difference between classes influences how often class distinctions will be removed.

\paragraph{Privacy versus utility}

When comparing PII-removal and DP-rewriting, we saw that the F1-scores approximating the utility have been far better when using PII-removal than when using DP-rewriting. However, this is not the case for privacy. We will discuss this further in the following.

In general, we know that one of the key points of DP-rewriting is that we can control the privacy risk, while in PII-removal there are no privacy guarantees. By setting the $\epsilon$ value in DP-rewriting, we can essentially set an upper boundary for the probability of a privacy leakage. For PII-removal, there are no guarantees at all. If we want to ensure that there are no privacy leakages, we would need to check every rewritten text for potential privacy leakages. Of course, this is unfeasible for larger datasets. Therefore, in practice, one would try to improve the PII-removal as much as possible and then hope that there are no privacy leakages, without knowing how high the risk for such a leakage exactly is. 

We will discuss what this means for our data in the following. For this, we will look at how many words of the input text have been changed or replaced. Of course, changing the wording is required but not sufficient to guarantee privacy. However, measuring the exact level of privacy preservation is hard and looking at the number of changed and replaced words is enough to give us a rough impression of how this minimal requirement was fulfilled on our data.

\begin{figure*}
    \centering
    \includegraphics[width=\textwidth]{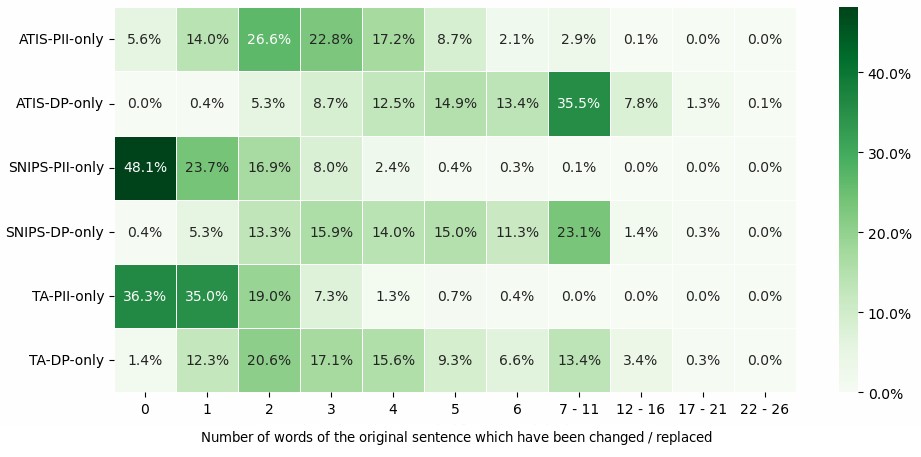}
    \caption{Distribution of the number of data points by the number of words from the original sentence that have been changed / replaced. E.g. 48.0\% in SNIPS-PII-only and 0 means that for 48.0\% of the data points of the SNIPS corpus the PII-only version contains the same words as the original sentence. Attention: look at the x-axis closely. There is a single column for each of the values from zero to six. Starting at value seven, we summed up the fractions for five values per column.}
    \label{fig:copied_words_distribution}
\end{figure*}

The heatmap in Figure~\ref{fig:copied_words_distribution} shows the results of this analysis per corpus and rewriting method. For a better understanding of this heatmap, we will explain one row as an example. The first row represents the PII-only version of the ATIS corpus. The value of the first column (``0'') is 5.6\%. This means, that for 5.6\% of all data points of the ATIS corpus, zero (``0'') words of the original sentence have been replaced or changed during PII-removal. So all words of the original sentence were copied into the PII-only version. In the next column (``1''), the value is 14\%, which means for 14\% of all data points of the ATIS corpus there is one word of the original sentence which has been changed or replaced during PII-removal. It continues like this for the next few columns. Then there is a column called ``7 - 11'', which is an aggregated column. The value 2.9\% tells us that for 2.9\% of all data points of the ATIS corpus between seven and eleven words of the original sentence have been replaced in the PII-only version of that sentence. The following columns are to be understood the same way.

In general, we see that with PII-only fewer words have been replaced than with DP-only. Especially for the SNIPS and TA data, there were many sentences which have not been changed at all (SNIPS: 48.1\%, TA: 36.3\%). Privacy preservation completely failed for these data points. Additionally, the amount of sentences where only a few words have been changed is also quite high when using PII-only. The privacy preservation to expect from those few changes might also be quite low. Therefore, the minimal requirement for privacy preservation, to change and/or replace words, has been fulfilled far better by DP-only than by PII-only.

However, there is one exception, where PII-only did not work that badly regarding privacy preservation. In the ATIS corpus, we see that in general a lot more words have been replaced by PII-only than in the other corpora. This is due to the fact that there are many easy-to-detect and therefore easy-to-replace PIIs in ATIS. Locations, dates and times can be detected quite well and ATIS is full of locations, dates and times. In SNIPS and TA, there are in general fewer of these easy-to-detect PII and additionally, the often uncommon sentence structures in SNIPS and TA make it harder to detect them. Therefore, PII-only was able to detect and therefore replace more PIIs in the ATIS corpus than in the SNIPS and TA corpora.

Nevertheless, there were also a noticeable number of examples in which PII-only failed in the ATIS corpus. For example, the original sentence ``what flights from indianapolis to memphis'' has been changed to ``what flights from <LOCATION> to memphis'' by PII-only. Obviously, ``memphis'' has not been recognized as a location. There are more examples like this. While one could try to further improve the PII-removal, as discussed before, there is no way to know how well privacy is preserved if you do not either have data in which all PII are labeled or manually check all texts.

All in all, we see that the performance of PII-only regarding privacy preservation is very domain specific. In general, PII-only replaces fewer words than DP-only. Furthermore, with DP-only one can set the upper bound for the probability of a privacy leakage, while with PII-only you do not have any guarantees.

\section{Conclusion and future work} 
In this work, we explored the effects of applying different privacy-preserving rewriting methods on textual data used for crowdsourcing. We compared PII-removal and DP-rewriting as well as a combination of both regarding utility and privacy.

PII-removal turned out to be a simple-to-implement approach that effects the utility least. However, there are no privacy guarantees given. DP-removal decreases the utility while at the same time giving privacy guarantees and decreasing the risk of privacy leakages. The utility decrease is highly dependent on the type of task and data. 

Therefore, based on our findings, we can give the following recommendations when using DP-rewriting. First, it is important to ensure that the pretraining data has an appropriate size based on the corpus and task.
The higher the similarity between classes as well as the diversity in sentence structures and wording of the corpus is, the more pretraining data is needed. Second, pretraining data should in the best case be balanced.



Future work should focus on overcoming the current shortcomings of current DP text rewriting approaches, namely the need to use very high values for $\epsilon$ which result in very low privacy guarantees.

\section{Limitations and ethical impact}

Regarding the corpora, important limitations are that we only requested annotations
for three corpora of which at least two had quite simple tasks. With only three corpora
there is not that much diversity in the selected corpora so that generalizing our results to
other corpora is harder. Therefore, we originally aimed to experiment with more corpora.
However, DP-rewriting did not work well enough for half of the originally chosen corpora, therefore we
needed to exclude them. While the low number of corpora was one problem, another
problem was that the selected corpora and their corresponding tasks were mostly quite
simple. We were able to identify a very small set of what we called indicator words for
ATIS and SNIPS and a larger set of indicator words for TripAdvisor. Probably, automatic
labeling dependent on these indicator words might have already worked quite well. This
makes generalization to more complex tasks even harder.

Apart from the used corpora, also the used rewriting methods cause some limitations.
First, we needed to use very high $\epsilon$-values for DP-rewriting in order to guarantee some
basic utility. However, these high $\epsilon$-values might not guarantee sufficient privacy in most
scenarios. Second, also PII-removal causes some limitations. PII-removal is very domain
dependent. Therefore, transferring our results to other domains is difficult. Furthermore,
PII-removal did not work that well for SNIPS and TripAdvisor, since in these corpora PII were harder to identify. Therefore, there were many cases were PII-removal just resulted
in copying the input text which resulted in zero privacy.

\bibliography{anthology,custom}
\bibliographystyle{acl_natbib}

\appendix

\section{Used Indicator Words}
\label{sec:appendix}
For ATIS and SNIPS, we used a manually curated list of indicator words. These words indicate that a text belongs to the target class. All used indicator words / phrases can be seen in Table~\ref{tab:indicator_wordlist}.

\begin{table}[ht]
    \centering
    \begin{tabularx}{\columnwidth}{c|X}\toprule
        Corpus & target class indicator words \\ \midrule
        ATIS & airfare, cheapest, cost, fare, fares, how much, price  \\
        SNIPS & add, playlist \\ \bottomrule
    \end{tabularx}
    \caption{Used target class indicator words for ATIS and SNIPS.}
    \label{tab:indicator_wordlist}
\end{table}

For TripAdvisor, the absence of negatively connoted words indicated that a review was positive. Therefore, we created a list of negatively connoted words. We used the lexicon of VADER \cite{Hutto14} to determine positively / negatively connoted words. We only included words where the sentiment was clear. Therefore, we excluded all words where adding or subtracting the doubled standard deviation from the polarity value would change the polarity.

\section{Example HIT}

\label{sec:example_HIT}
\begin{figure*}[t]
    \centering
    \includegraphics[width=\textwidth]{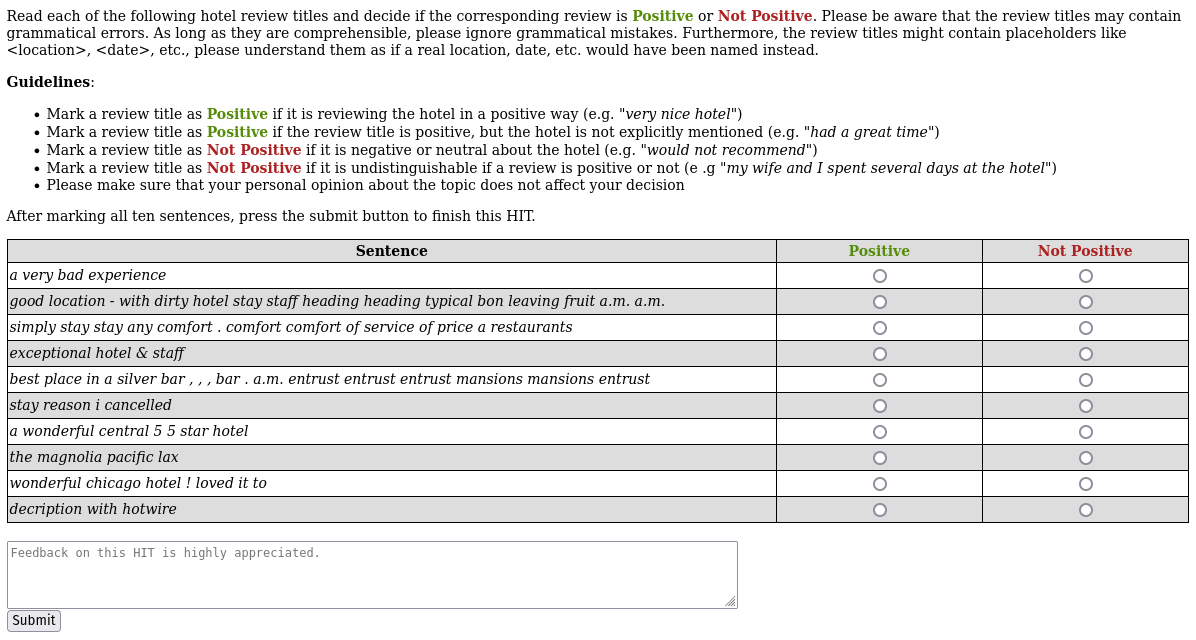}
    \caption{Screenshot of an example HIT. This HIT is filled with DP-only data of the TA corpus.}
    \label{fig:example_HIT}
\end{figure*}

\end{document}